\documentclass[a4paper]{svproc}

\usepackage{url}

\usepackage{amsmath,amssymb,amsfonts}
\usepackage{makecell}
\usepackage{threeparttable}
\usepackage[colorlinks, linkcolor=blue]{hyperref}
\usepackage{xcolor}
\usepackage{graphicx} 
\makeatletter

\newcommand{\Rmnum}[1]{\expandafter\@slowromancap\romannumeral #1@}
\makeatother
\begin{document}
\mainmatter              
\title{Grasping by Hanging: a Learning-Free Grasping Detection Method for Previously Unseen Objects}
%
%
\author{Wanze Li\inst{1} \and Wan Su\inst{1} \and Gregory S. Chirikjian\inst{1, 2}}

\institute{Mechanical Engineering Department, National University of Singapore, Singapore\\
\email{li\_wanze@u.nus.edu}, \email{suwan@u.nus.edu}\\ 
\and
Mechanical Engineering Department, University of Delaware, Newark, Delaware, United States \\
\email{gchirik@udel.edu}}

\maketitle              

\begin{abstract}
This paper proposes a novel learning-free three-stage method that predicts grasping poses, enabling robots to pick up and transfer previously unseen objects. 
Our method first identifies potential structures that can afford the action of hanging by analyzing the hanging mechanics and geometric properties. 
Then 6D poses are detected for a parallel gripper retrofitted with an extending bars, which when closed form loops to hook each hangable structure. 
Finally, an evaluation policy qualities and rank grasp candidates for execution attempts. 
Compared to the traditional physical model-based and deep learning-based methods, our approach is closer to the human's natural action of grasping unknown objects. 
And it also eliminates the need for a vast amount of training data. 
To evaluate the effectiveness of the proposed method, we conducted experiments with a real robot. 
Experimental results indicate that the grasping accuracy and stability are significantly higher than the state-of-the-art learning-based method, especially for thin and flat objects. 

\end{abstract}
\section{Introduction}

Object grasping is a common and crucial operation in industrial, agricultural and domestic environments. 
Therefore, enabling robots to grasp various objects is a long-standing topic of interest in robotics.
Recently, with the rapid development of machine learning, data-driven methods have dominated the field of robotic grasping \cite{1}, \cite{2}, \cite{3}, \cite{4}, \cite{5}.
These learning-based methods significantly improved the grasping performance. 
However, a large amount of high-quality data is necessary to train these methods, which is hard to obtain and requires substantial computing resources. 
Moreover, data-driven methods tend to perform poorly in objects with small thickness \cite{6}, \cite{7}. 
\begin{figure}[htbp]
\centerline{\includegraphics[width=5in]{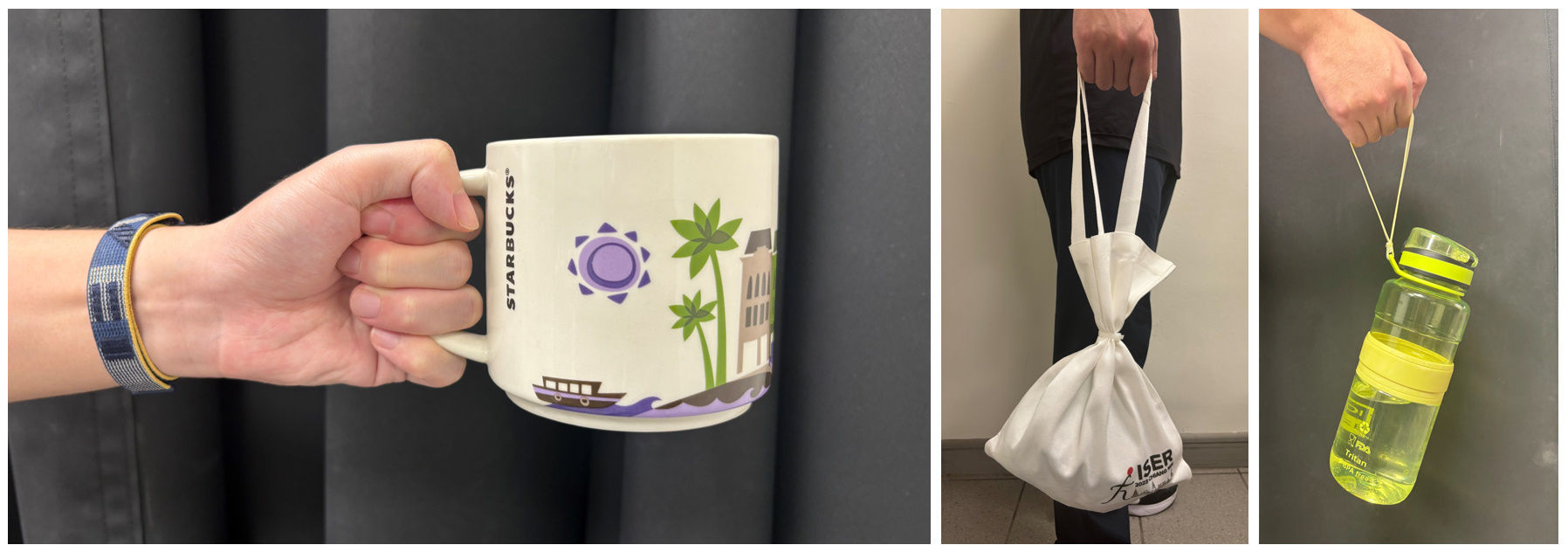}}
\caption{Examples of a human picking objects with convenient parts for grasping. }
\label{intro}
\end{figure}
On the other hand, when humans face previously unseen objects, we can identify and grasp specific parts of objects that are convenient for grasping, such as the handle of cup and bag shown in Fig.\,\ref{intro}. 
To adapt the philosophy of human perception and action in object grasping, we propose a novel framework for grasping pose detection of a parallel jaw gripper in this paper. 

The idea of grasping objects with specific structures has been utilized in some previous works \cite{8}, \cite{9}, \cite{10}, \cite{11}, \cite{12}. 
For instance, in \cite{9}, Pokorny et al. developed a topological approach to detect the holes in objects and generate grasping poses with these holes. 
However, their method detects holes in objects by identifying the topologically closed loops, which have very high requirements for the quality of perception. 
Moreover, some unenclosed great candidates for grasping, like the hook of a clothes hanger, cannot be detected. 
Another related research topic is robot caging \cite{13}, \cite{14}, \cite{15}, \cite{16}, \cite{17}, which is a kind of grasping method without considering mechanical properties \cite{11}. 
Nevertheless, most of the robot caging research focuses on grasping with multi-finger or multi-freedom of each finger, which requires more complex control methods as well as higher economic costs.  

\begin{figure*}[htbp]
\centerline{\includegraphics[width=5in]{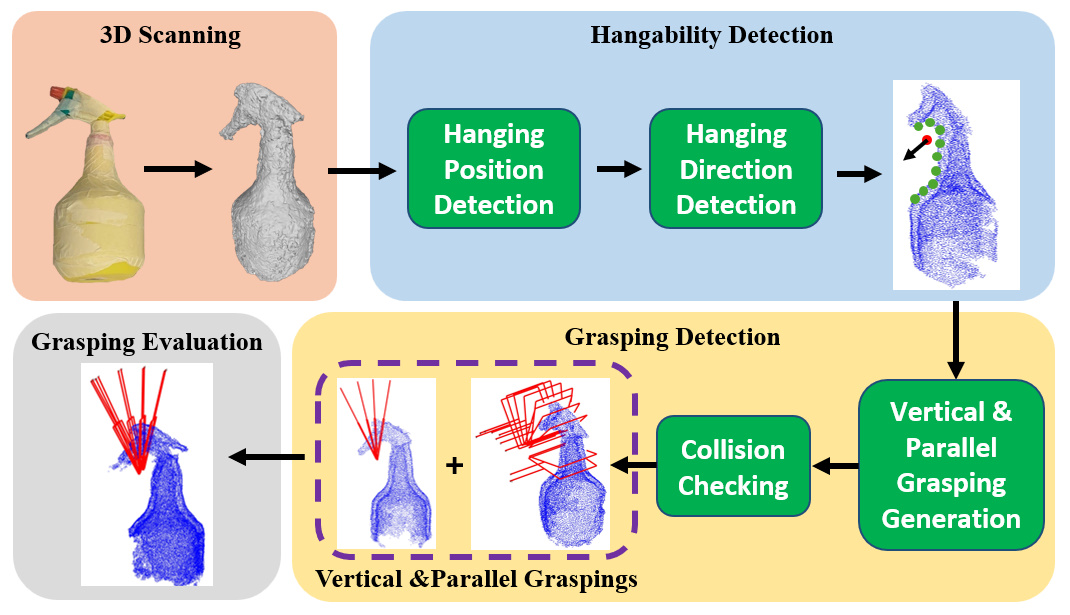}}
\caption{The workflow of the proposed method. The entire procedure includes 3D scanning, hangability detection, grasping detection and grasping evaluation. The hanging position and corresponding hanging direction are marked as red point and black arrow, respectively. Grasping poses are denoted as red lines. }
\label{workflow}
\end{figure*}

From another perspective, the grasping shown in Fig.\,\ref{intro} can be regarded as the objects hanging supported by human hands. 
Therefore, the grasping poses can be generated by detecting the parts of the object that can afford hanging and corresponding hanging information. 
Based on this, this paper proposes a three-stage 6D grasping pose detection method (Fig.\,\ref{workflow}). 
Firstly, the algorithm takes the mesh of the object as input and identifies the suitable hanging positions and directions based on our previous work \cite{18}. 
Hanging positions and directions are represented as 3D key points and key vectors, respectively.   
Subsequently, a series of collision-free grasping poses are generated for each hangable structure according to the hanging position and direction. 
In this last stage, all grasping poses are evaluated and ranked to determine the most suitable grasping pose for execution. 
Experiments are conducted on isolated objects in the real world to validate our method. 
The experimental results show that our method substantially outperforms the state-of-the-art in terms of performance. 

The main contributions of this paper include: 
(1) a learning-free object grasping detection method inspired by human grasping and objects hanging, 
(2) a straightforward, simple, yet effective evaluation method to evaluate
the quality of each grasping candidate, and 
(3) real-world validation of the proposed method by grasping previously unseen daily objects. 

\section{Method}

This section introduces the details of the proposed grasping-by-hanging (GbH) method. 
Firstly, the structure of the parallel gripper we used in this work is introduced in \ref{gripper_design}. 
Then we describe the procedure of detecting the hanging positions and directions in \ref{Hangability}. 
Subsequently, \ref{Generation} explains how to generate the grasping poses based on the result of \ref{Hangability}. 
Finally, grasping candidates are evaluated in \ref{Evaluation}. 
The workflow of the entire grasping detection procedure is displayed in Fig.\,\ref{workflow}. 
\vspace{-1mm}
\begin{figure}[htbp]
\centerline{\includegraphics[width=5in]{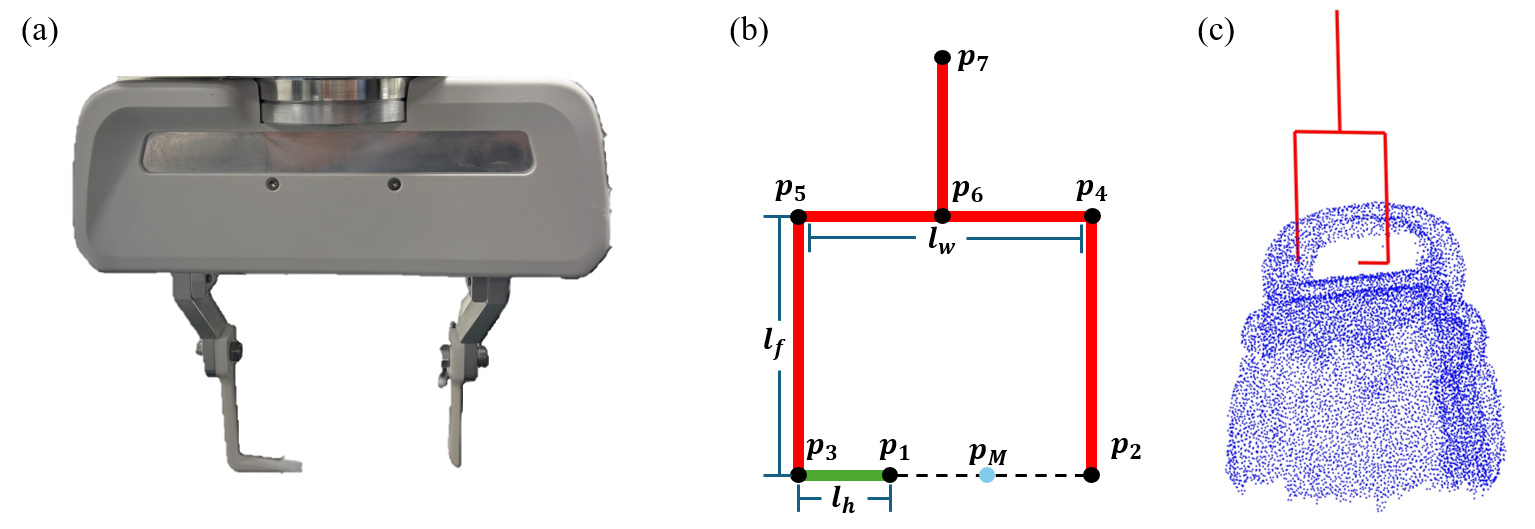}}
\caption{(a) The image of the gripper used in this work. (b) The simplified model of the gripper. The length of the finger is denoted as $l_f$, the maximum width of the gripper is denoted as $l_w$ and the length of the horizontal rod is $l_h$. We also denote endpoints of fingers and hand as $\mathbf{p}_1$ to $\mathbf{p}_7$. The middle point between the endpoint of the horizontal rod and the finger on the other side is denoted as $\mathbf{p}_M$. (c) An example of the gripper at the grasping pose. }
\label{gripper}
\end{figure}

\subsection{Parallel Jaw Gripper} \label{gripper_design}
As shown in Fig.\,\ref{gripper}a, the gripper used in this paper is modified from Franka Emika hand with parallel jaws. 
Two fingers are extended by one straight rod and one L-shaped rod with the same length so that the object can be caged when the gripper is closed. 
This paper denotes the finger length as $l_f$, the maximum width of the gripper as $l_w$ and the length of the horizontal part of the L-shaped rod (marked as the green line in Fig.\,\ref{gripper}b) as $l_h$. 
Moreover, endpoints of the gripper are denoted as $\mathbf{p}_1$ to $\mathbf{p}_7$. 
Finally, the middle point between the endpoint of the horizontal rod ($\mathbf{p}_1$) and the fingertip on the other side ($\mathbf{p}_2$) is denoted as $\mathbf{p}_M$. 
In this work, $l_w = l_f = 8cm$ and $l_h = 3cm$. 

\begin{figure}[htbp]
\centerline{\includegraphics[width=5in]{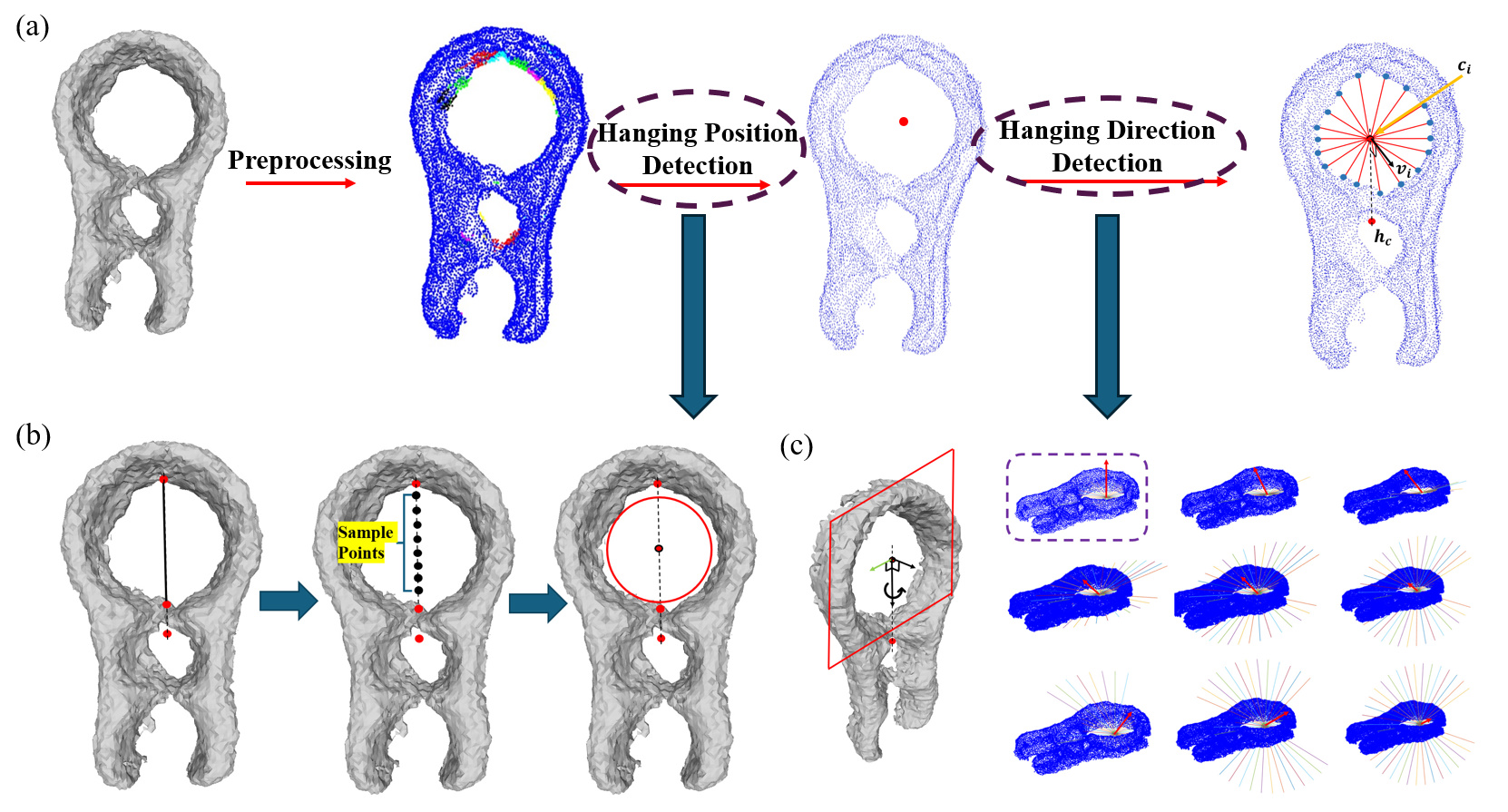}}
\caption{(a) The workflow of the hangability detection. The rightmost sub-figure shows the final result. The hanging position is represented as $\mathbf{c}_i$ and the corresponding hanging direction is denoted as $\mathbf{v}_i$. $\mathbf{h}_c$ represents the center of mess. Red lines denote the casting rays in the hanging direction detection and blue points denote contact points between rays and the object. (b) The diagram of the hanging position detection. (c) The diagram of the hanging direction detection.  }
\label{hang}
\end{figure}

\subsection{Hangability Detection} \label{Hangability}
The goal of hangability detection is to find all possible structures in the object that could be used to hang it steadily. 
Each hangable structure is represented as a hanging position and a hanging direction. 
Following our previous work in \cite{18}, this stage takes the mesh of the object as input and samples a point cloud from it by poisson disk sampler \cite{19}. 
The potential contact points for hanging are calculated by clustering points with normal vectors pointing to the center of the mess (CoM) based on the force analysis in \Rmnum{2}-A. 
Next, as shown in Fig.\,\ref{hang}b, the point with the most free space around it on the line segment between the potential contact point and the center of the mess (CoM) is selected as the hanging position. 
The procedure of detecting corresponding hanging directions is displayed in Fig.\,\ref{hang}c. 
For each hanging position, the corresponding hanging direction is determined by casting rays within different planes and choosing the normal vector of the plane with the most number of rays that intersect with the object as the hanging direction. 
Since for now in this work, only the `through hole' case is considered, two rays are cast at each hanging position along the positive and negative hanging directions. 
The hanging position/direction pair is kept for the following stages only if both rays have no intersection with the object mesh. 
The flowchart of this stage is shown in Fig.\,\ref{hang} and more details about the hangability detection can be found in in \cite{18}. 

Eventually, the following information is acquired after the hangability detection: 
(1) hanging positions: $C=\{\mathbf{c}_i\}_{i=1}^N$, where $N$ is the number of hanging positions; 
(2) hanging directions: $V=\{\mathbf{v}_i\}_{i=1}^N$, $v_i$ is an unit vector; 
(3) contact points on ray casting detection: $H=\{H_i\}_{i=1}^N$, where $H_i=\{\mathbf{h}_i^j\}_{j=1}^K$ represents the set of $K$ contact points between rays and the mesh for the $ith$ hanging position;  
(4) the proportion of rays intersecting with objects at each hanging position: $m=\{m_i\}_{i=1}^N$, which represents the `completeness' of the object encircling the hanging position and 
(5) the average direction of rays that don't intersect with the object: $A=\{\mathbf{a}_i\}_{i=1}^N$, $\mathbf{a}_i$ is the mean of directions of rays that miss the object during the $ith$ hanging direction detection. 
Examples of the hangability detection result are displayed in Fig.\,\ref{grasp_generation}a.

\subsection{Grasping Generation} \label{Generation}
The next stage involves synthesizing available grasping poses with the result of hangability detection. 
In general, two different kinds of grasping poses are synthesized by aligning the horizontal rod ($\overline{\mathbf{p}_1\mathbf{p}_3}$ in Fig.\,\ref{gripper}b) and the straight finger ($\overline{\mathbf{p}_2\mathbf{p}_4}$) with the hanging direction. 
In this work, these two kinds of grasping are named parallel grasping and vertical grasping, respectively. 
As a result, the gripper plays the role of supporting item in \cite{18} at these poses so that the object can be picked up and transported by the gripper safely. 
This paper denotes the unit vector $\frac{\overrightarrow{\mathbf{p}_7\mathbf{p}_6}}{|\overrightarrow{\mathbf{p}_7\mathbf{p}_6}|}$ as $\mathbf{n}_1$ and $\frac{\overrightarrow{\mathbf{p}_3\mathbf{p}_1}}{|\overrightarrow{\mathbf{p}_3\mathbf{p}_1}|}$ as $\mathbf{n}_2$. 

\begin{figure}[htbp]
\centerline{\includegraphics[width=5in]{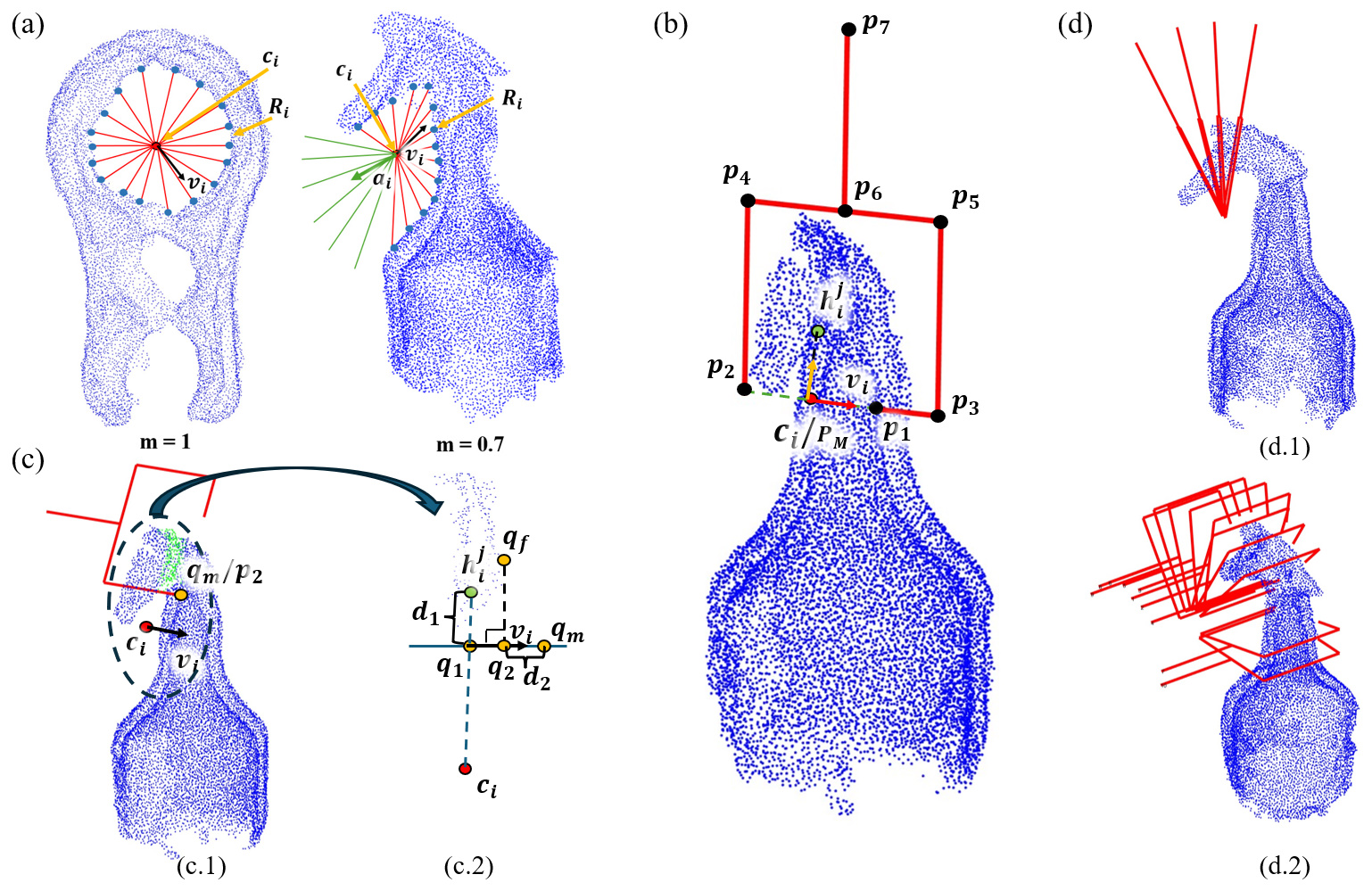}}
\caption{(a) The image to show the result of hangability detection of two objects. Hanging centers are marked as red points, and hanging directions are denoted as black arrows. Red lines and green lines represent the rays that with or without intersect with the object, respectively. The intersection points between the rays and the mesh are denoted as blue dots. (b) An example of a parallel grasping. In this case, $\overrightarrow{\mathbf{p}_7\mathbf{p}_6}$ is parallel with $\overrightarrow{\mathbf{h}_i^j\mathbf{c}_i}$, $\overrightarrow{\mathbf{p}_1\mathbf{p}_3}$ is parallel with $\mathbf{v}_i$, the $\mathbf{p}_M$ coincides with the $\mathbf{c}_i$. (c) An example of one vertical grasping. In this case, $\overrightarrow{\mathbf{p}_3\mathbf{p}_1}$ is parallel with $\overrightarrow{\mathbf{h}_i^j\mathbf{c}_i}$, $\overrightarrow{\mathbf{p}_4\mathbf{p}_2}$ is parallel with $\mathbf{v}_i$, the $\mathbf{p}_2$ coincides with the $\mathbf{q}_m$. Points caged by the gripper are marked as green in c.1 and c.2 illustrating the local details of determining $\mathbf{q}_m$. (d) Collision-free parallel and vertical grasping poses after the grasping generation are displayed in d.1 and d.2 respectively.}
\label{grasp_generation}
\end{figure}

\textbf{Parallel Grasping: }For each hanging direction $\mathbf{v}_i$, the horizontal rod ($\mathbf{n}_1$) is first aligned with $\mathbf{v}_i$ and $-\mathbf{v}_i$ to generate two groups of parallel grasping poses. 
Then for each contact point $\mathbf{h}_i^j \in H_i$, a grasping pose can be determined by rotating the gripper about $v_i$ until $\mathbf{n}_1$ parallel with the vector $\overrightarrow{\mathbf{h}_i^j\mathbf{c}_i}$. 
Finally, the middle point between the endpoint of the horizontal rod and the straight finger ($\mathbf{p}_M$ in Fig.\,\ref{gripper}) is translated to overlap with the hanging position $\mathbf{c}_i$. 
As a result, for each hanging position $\mathbf{c}_i$, $K$ grasping poses are generated, where $K$ is the number of elements in $H_i$. 
Fig.\,\ref{grasp_generation}b illustrates an example of grasping poses synthesized for one hanging position. 

\textbf{Vertical Grasping: }For vertical grasping generation, similar to the case of parallel grasping, potential grasping poses are sampled by rotating the gripper and aligning $\mathbf{n}_1$ with each vector $\overrightarrow{\mathbf{h}_i^j\mathbf{c}_i}$. 
Then the straight finger ($\overline{\mathbf{p}_2\mathbf{p}_4}$ in Fig.\,\ref{gripper}b) is aligned with the hanging direction $\mathbf{v}_i$. 
Since the object might be lying on the ground and pick along the hanging direction might cause the collision with the ground, the straight finger is aligned with the normal vector of the ground if $cos(\theta) > p_{\theta}$. 
Here $\theta$ represents the angle between the hanging direction and the ground normal vector, $p_{\theta}=0.95$ is a predefined threshold. 

Now the rotations of all vertical grasping poses are determined, the next step is determining the translation by matching the point $\mathbf{p}_2$ of Fig.\,\ref{gripper}b with a proper position $\mathbf{q}_m$ obtained from the hanging information. 
The procedure of determining $\mathbf{q}_m$ is illustrated in Fig.\,\ref{grasp_generation}c. 
Firstly, a point $\mathbf{q}_1$ is calculated by: 
\begin{equation}
  \mathbf{q}_1 = \mathbf{h}_i^j + d_1*\frac{\overrightarrow{\mathbf{h}_i^j\mathbf{c}_i}}{|\overrightarrow{\mathbf{h}_i^j\mathbf{c}_i}|}
\end{equation}
Then the object point cloud is sliced by the plane defined by the gripper to obtain the points that caged by the gripper when $\overline{\mathbf{p}_2\mathbf{p}_4}$ passes $\mathbf{q}_1$ (green points in Fig.\,\ref{grasp_generation}c.1). 
The set of these points is denoted as $Q$. 
Then the point in $Q$ that is farthest along $\mathbf{v}_i$ is selected and defined as $\mathbf{q}_f$: 
\begin{equation}
  \mathbf{q}_f = \mathop{\arg\max}\limits_{\mathbf{q}_i \in Q}((\mathbf{q}_i-\mathbf{q}_1)\cdot\mathbf{v}_i)
\end{equation}
And the projection of $\mathbf{q}_f$ on the line defined by $\mathbf{q}_1$ and $\mathbf{v}_1$ is denoted as $\mathbf{q}_2$ (labelled in  Fig.\,\ref{grasp_generation}c.2). 
Then, the matching point $\mathbf{q}_m$ is calculated by: 
\begin{equation}
  \mathbf{q}_m = \mathbf{q}_2 + d_2*\mathbf{v}_i
\end{equation}
to ensure that the object can be caged by the jaws once the gripper is closed. 
Eventually, the translation of vertical grasping is determined by matching the point $\mathbf{p}_2$ of the gripper with $\mathbf{q}_m$. 
In this step, $d_1$ and $d_2$ are predefined parameters. 

\textbf{Collision Checking: }To ensure safe and successful grasping, it is essential that predicted grasping poses are collision-free. 
In this work, collision checking is achieved by counting the number of points from the object's point cloud within the gripper model at the grasping pose. 
The number of these points is denoted as $N_c$ and only grasping poses that satisfy $N_c < p_c$ are kept for the following steps. 
Here, $p_c$ is a predefined threshold and selected as 10. 

The set of collision-free grasping poses obtained in the grasping generation stage is denoted as $G = \{g_i=(\mathbf{R}_i, \mathbf{t}_i)\}_{i=1}^M$, where $\mathbf{R}_i$ is the rotation matrix and $\mathbf{t}_i$ is the translation of the $ith$ grasping pose. 

\subsection{Grasping Evaluation} \label{Evaluation}
 The final stage of the proposed method aims to evaluate the quality of predicted grasping poses and select the most proper one. 
The grasping evaluation is implemented by calculating and ranking the value of a score function that considers the grasping direction and the completeness of the free space. 

\textbf{Grasping Direction: }As humans pick objects, the gripper moves in the anti-gravity direction once the object is grabbed. 
Therefore, when the grasping direction is close to the gravitational direction, it is easier for the robot to pick up the object. 
Moreover, approaching the object from the top down provides better avoidance of collision with the environment. 
The score of the grasping direction for a grasping pose $g_i=(\mathbf{R}_i, \mathbf{t}_i)$ is defined as: 
\begin{equation}
  S_{\alpha}:= exp(-\frac{\alpha^2}{\gamma_\alpha})
\end{equation}
where $\alpha \in [0, \pi]$ is the angle between the vector obtained by rotating $\mathbf{n}_1$ for $\mathbf{R}_i$ and the anti-gravitational direction. 
The grasping direction with a higher $S_\alpha$ value is closer to the anti-gravitational direction, meaning the grasping pose is safer and easier to reach. 

\textbf{Free Space Completeness: }When the contact point between the object and the gripper is close to the `open' area (green rays in Fig.\,\ref{grasp_generation}a), the object is more likely to fall out of the gripper. 
Therefore, the other term of the score function aims to motivate the robot to select the grasping pose that is farther away from the `open' area of the free space. 
The score for free space completeness is defined as: 
\begin{equation}
S_{\beta}:=\left\{
\begin{array}{l}
1, \qquad m_i =1 \\
exp(-\frac{\beta^2}{\gamma_\beta}), \qquad m_i < 1 \\
\end{array}
\right.
\end{equation}
where $\beta \in [0, \pi]$ is defined as: 
\begin{equation}
\beta:=\left\{
\begin{array}{l}
\arccos({\mathbf{a}_i\cdot(\mathbf{R}_i\mathbf{n}_1)}), \quad for \ parallel \ grasping \\
\arccos({\mathbf{a}_i\cdot(\mathbf{R}_i\mathbf{n}_2)}), \quad for \ vertical \ grasping \\
\end{array}
\right.
\end{equation}
Smaller $\beta$ means the gripper is farther away from the open area, which means the corresponding grasping pose is safer. 

Finally, the total score for $g_i$ is computed as: 
\begin{equation}
    S_{total} = m_i\times S_{\beta} \times S_{\alpha}
\end{equation}
In this work, we assign parameters $\gamma_\alpha$ and $\gamma_\beta$ as 0.04 and 2 respectively. 

\begin{figure}[htbp]
\centerline{\includegraphics[width=5in]{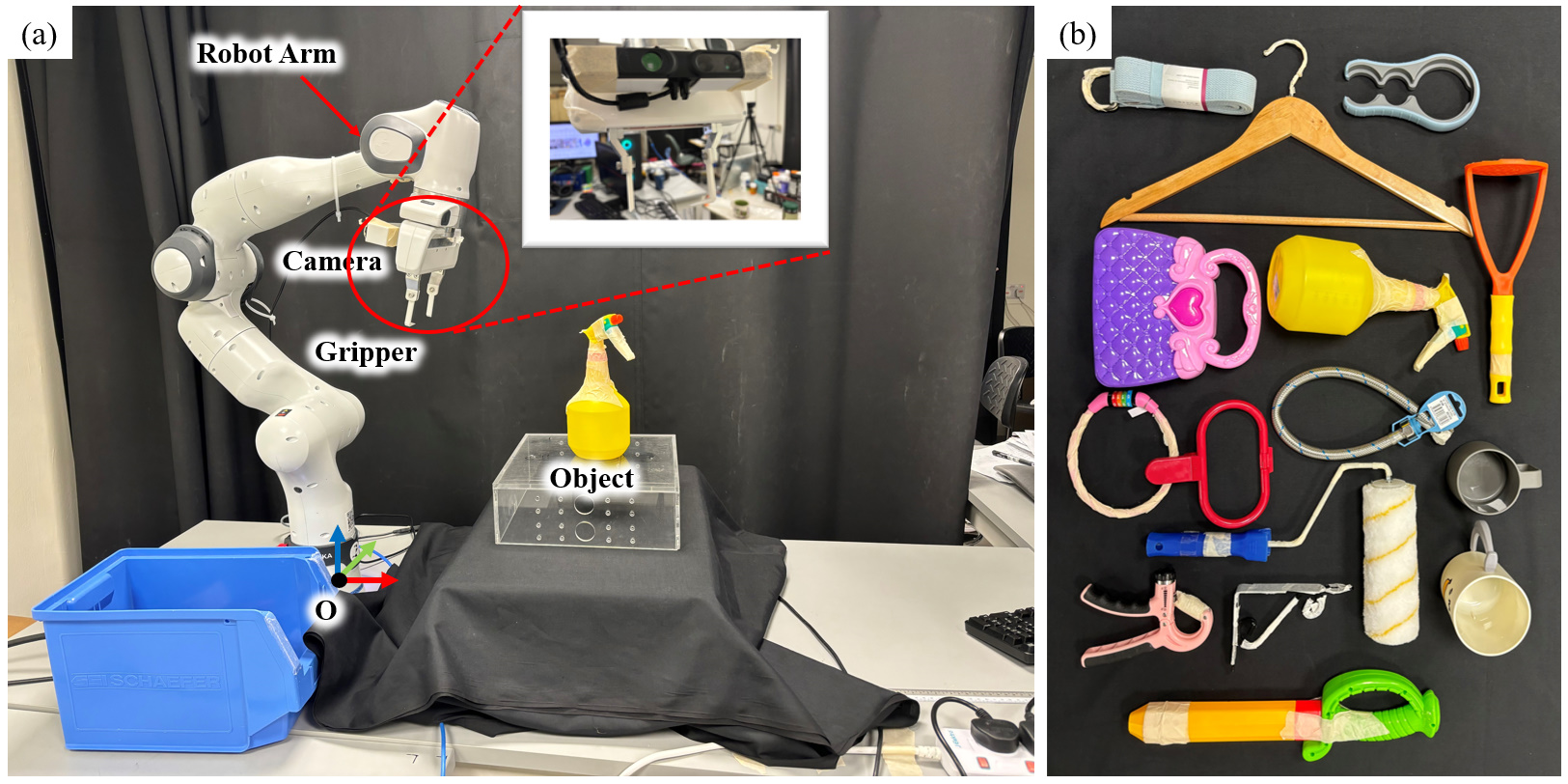}}
\caption{(a) The setup of the experiment. (b) Dataset used for experiments. The dataset contains 15 daily objects. }
\label{exp_setting}
\end{figure}

\section{Experiment}
\subsection{Experimental Setting}
To evaluate the performance of the proposed approach, we conduct real-world experiments with 15 objects (shown in Fig.\,\ref{exp_setting}b). 
These objects are previously unseen to the robot. 
As displayed in Fig.\,\ref{exp_setting}a, a Franka Emika robot arm mounted with a Primesense RGB-D camera is used for object scanning, grasping and manipulation. 
The gripper of the robot arm is modified as described in \ref{gripper_design} for the execution of the proposed method. 
In each trial, the mesh of the object is input to the algorithm for grasping pose prediction. 
Ten predicted poses with the highest $S_{total}$ value are selected and ranked from high to low. 
The robot executes the one with the highest score that an RRT-connect \cite{20} motion planner can find an available trajectory of it. 
If none of these ten poses are reachable, this trial is considered a failure. 

We conduct three groups of experiments: 
(1) grasping-by-hanging (GbH) with 10 viewpoints for 3D scanning; 
(2) grasping-by-hanging (GbH) with 1 viewpoint for 3D scanning; 
(3) GraspNet \cite{1} in our dataset
to compare the performance of GbH to the learning-based baseline and explore the impact of the scanning quality.  
In each experimental group, the robot attempts to grasp one of the objects in Fig.\,\ref{exp_setting}b, with 5 trails for each object. 
After grasping, the robot transports the object to a specific position within the workspace and releases it. 
One trail is considered as successful only if the object is picked up and transported without falling during the entire procedure. 
In the experiment, we use TSDF Fusion \cite{21} for mesh reconstruction and Open3D \cite{22} to process meshes and point clouds. 
But our method has no limitation to the 3D reconstruction step, any method, like \cite{23} can be used. 
Algorithms are implemented in Python on a computer equipped with AMD Ryzen™ 9 5950X (3.4GHz).
In the experiment with GbH, we select $d_1=0.01, d_2=0$ for the experiment with 10 scanning viewpoints and $d_1=0.01, d_2=0.005$ for a single viewpoint to compensate for the incompleteness of the model. 
For the sake of fairness, when experimenting with baseline, the gripper is mounted with two straight fingers of the same length as the one used for GbH. 

\begin{figure}[htbp]
\centerline{\includegraphics[width=5in]{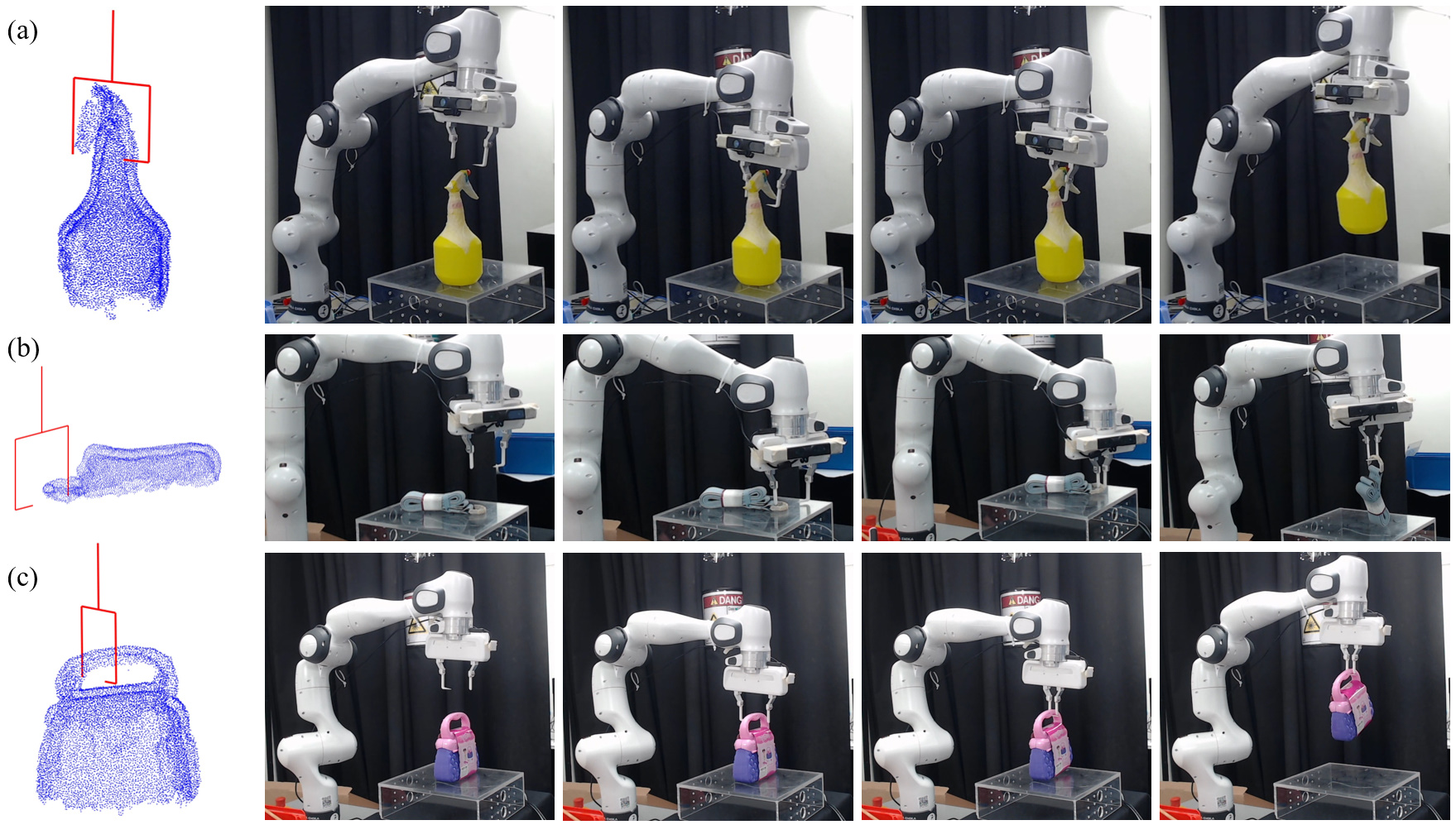}}
\caption{Examples of the predicted grasping poses by GbH and the robotic execution. }
\label{exp_example}
\end{figure}

\subsection{Results and Discussion}
\begin{table}[h]
    \setlength{\tabcolsep}{6pt} 
    \renewcommand{\arraystretch}{1.2}
    \centering
    \caption{Grasping Accuracy of Real-World Experiments}
    \begin{tabular}{cccccc}
    \hline
     Method & Viwepoints &Total & \makecell[c]{Success} & \makecell[c]{Fail} & \makecell[c]{Rate of \\ Success} \\
    \hline 
    GraspNet \cite{1} & $1$ & $75$ & $23$ & $52$ & $30.7\%$ \\
    GbH & $1$ & $75$ & $54$ & $21$ & $72\%$  \\
    GbH & $10$ & $75$ & $68$ & $7$ & $90.7\%$  \\
    \hline
    \end{tabular}
    \label{tab:exp_res}
\end{table}
\textbf{Result: }Experimental result is displayed in the Table \ref{tab:exp_res}. 
The grasping success rate of the proposed method significantly outperforms the baseline method ($30.7\%$) with both single viewpoint ($72\%$) and 10 viewpoints ($90.7\%$) for scanning. 
Even with the same input, the success rate of our method is still more than twice as much as the baseline's. 
Moreover, our method does not require any training data but the baseline method is trained on a dataset with 97,280 RGB-D images and more than 1 billion grasping poses. 
Some examples of grasping poses predicted by GbH and the following execution are displayed in Fig.\,\ref{exp_example}. 

\textbf{Discussion: }The advantages of the proposed method are particularly evident when grasping thin and flat objects. 
For instance, the baseline usually cannot find any grasping pose for the clothes hanger, but our method achieves a $90\%$ rate of success in grasping the clothes hanger. 
For some other objects, like the cup and the sprinkling can, the baseline method tends to predict grasping poses on the body. 
The diameter of the body might be close to or even larger than the maximum width of the gripper, which makes grasping extremely difficult, even impossible.
On the contrary, our method clings to picking the object with the handle and naturally achieves more success in these objects.  
In addition, only one failure over 150 attempts of GbH due to the object falling off during movement, which means the grasping stability of the proposed method is high. 
When the gripper is closed at the pose predicted by our method, the object is caged by the straight finger, L-shaped finger and the gripper base. 
As a result, the likelihood of objects falling off from the gripper during transportation is highly reduced, especially when the value of $m$ in \ref{Hangability} is 1, meaning the object is grabbed in a `loop-on-loop' manner (like the case in Fig.\,\ref{exp_example}b and c). 

\textbf{Failure Analysis: }When the object is fully scanned, the main reason for the grasping failure is that no hanging position/direction is detected. 
This issue can be improved by optimizing the parameter settings of the hangability detection. 
Another issue worth noticing is the sensitivity to scanning. 
When the scanning quality is low, some parts in the object that can afford hanging, like the cup handle, become incomplete in the mesh. 
As a result, hangings and grasping poses with these parts can not be detected. 
Besides that, when the object is only partially perceived, some structures might be incorrectly identified as hangable. 
For example, when the cup is only scanned from one viewpoint, some holes occur on the cup wall, which can mislead the algorithm into predicting the poses of grasping the object with these holes. 
Finally, incomplete meshes and point clouds also reduce the accuracy of collision checking. 

\textbf{Limitations: }Although the experimental results are encouraging, our method comes with certain limitations. 
The main limitation is that currently GbH method only considers grasping objects with hangable structures. 
Although it has already covered a wide range of objects, there is still room for further improvement in the universality of the method.
For instance, we plan to integrate GbH with other parallel jaw grasping detection methods, either learning-free like \cite{10} or data-driven like \cite{24} shortly. 
A revolute joint will also be added between the $\overline{\mathbf{p}_5\mathbf{p}_3}$ and $\overline{\mathbf{p}_3\mathbf{p}_1}$ so that the horizontal bar can be retracted or released.
Therefore, once the object is considered hangable, the robot releases the horizontal bar and grasps the object by hanging. 
When no hangable structure is detected, the robot can retract the bar and grasp the object with regular methods. 
Another major limitation is that our method is still sensitive to the quality of scanning. 
To solve this issue, we plan to use some shape completion methods like the Gaussian process shape completion method used in \cite{25}. 

\section{Conclusion}
This paper proposes a learning-free framework for object grasping with a modified parallel jaw gripper. 
Our method first detects the structures in the object that can afford hanging and represents each structure as a hanging direction and a hanging position. 
Then potential grasping candidates are generated according to the hangability information.  
Finally, feasible grasping poses are determined after the collision-checking and evaluation. 
Experiments are conducted in the real-world environment to validate the proposed method and the result shows that the performance of our method is significantly higher than the baseline method. 
Future work will concentrate on integrating regular grasping methods and shape completion algorithms into our framework to increase universality and stability. 

\section{Acknowledgements}
This work was supported by the National Research Foundation, Singapore, under its Medium Sized Centre Programme - Centre for Advanced Robotics Technology Innovation (CARTIN), subaward A-0009428-08-00, and AME Programmatic Fund Project MARIO A-0008449-01-00. The authors thank Ceng Zhang in NUS for the help in figures making.
%
%

\end{document}